\documentclass[conference]{IEEEtran}
\makeatletter
\@ifundefined{pdfobjcompresslevel}{}{\pdfobjcompresslevel=3}
\makeatother
\IEEEoverridecommandlockouts

\usepackage{cite}
\usepackage{amsmath,amssymb,amsfonts}
\usepackage{algorithmic}
\usepackage{graphicx}
\usepackage{textcomp}
\usepackage{xcolor}
\def\BibTeX{{\rm B\kern-.05em{\sc i\kern-.025em b}\kern-.08em
    T\kern-.1667em\lower.7ex\hbox{E}\kern-.125emX}}

\usepackage[inline]{enumitem}
\usepackage{marvosym}
\usepackage{multirow}
\usepackage[caption=false, font=footnotesize]{subfig}
\usepackage{booktabs}
\usepackage[bookmarks=false, pdfborder={0 0 0}, hidelinks]{hyperref}

\makeatletter 
\newcommand{\linebreakand}{%
  \end{@IEEEauthorhalign}
  \hfill\mbox{}\par
  \mbox{}\hfill\begin{@IEEEauthorhalign}
}
\makeatother 

\usepackage{fontspec}

\usepackage{polyglossia}
\setmainlanguage{english}
\setotherlanguages{hindi,bengali,thai}
\newfontfamily\devanagarifont[Script=Devanagari, Path=./]{NotoSerifDevanagari-Regular.ttf}

\newfontfamily\bengalifont[Path=./, BoldFont={Kalpurush.ttf}, BoldItalicFont={Kalpurush.ttf}]{Kalpurush.ttf}

  \newfontfamily\thaifont[Script=Thai, Path=./]{NotoSerifThai-Regular.ttf}
  \DeclareRobustCommand{\nobreakspace}{\leavevmode\nobreak\ }

\addto\captionsenglish{%
  \renewcommand{\figurename}{Fig.}%
  }
  
\begin{document}

\title{When Meaning Isn't Literal: Exploring Idiomatic Meaning Across Languages and Modalities}

\author{\IEEEauthorblockN{
1\textsuperscript{st} Sarmistha Das\textsuperscript{\textdagger}, 2\textsuperscript{nd} Shreyas Guha\textsuperscript{\textdagger}, 3\textsuperscript{rd} Suvrayan Bandyopadhyay\textsuperscript{\textdagger}}
\IEEEauthorblockA{\textit{Department of Computer Science and Engineering} \\
\textit{Indian Institute of Technology Patna}\\
Patna, India\\
\{sarmistha\_2221cs21, 2201cb58\_shreyas, suvrayan\_2301cs89\}@iitp.ac.in}\thanks{\textsuperscript{\textdagger}Equal Contribution}
\linebreakand
\IEEEauthorblockN{4\textsuperscript{th} Salisa Phosit, 5\textsuperscript{th} Kitsuchart Pasupa\textsuperscript{*}}
\IEEEauthorblockA{\textit{School of Information Technology} \\
\textit{King Mongkut's Institute of Technology Ladkrabang}\\
Bangkok, Thailand\\
67076055@kmitl.ac.th, \textsuperscript{*}kitsuchart@it.kmitl.ac.th}\thanks{\textsuperscript{*}Corresponding Author}
\and
\IEEEauthorblockN{6\textsuperscript{th} Sriparna Saha}
\IEEEauthorblockA{\textit{Department of Computer Science and Engineering} \\
\textit{Indian Institute of Technology Patna}\\
Patna, India\\
sriparna@iitp.ac.in}
}

\maketitle

\begin{abstract}
Idiomatic reasoning deeply intertwined with metaphor and culture remains a blind spot for contemporary language models, whose progress skews toward surface-level lexical and semantic cues. For instance, the Bengali idiom \begin{bengali}আঙ্গুর ফল টক\end{bengali} (\textit{angur fol tok}, ``grapes are sour''): it encodes denial-driven rationalization, yet Na\"ive models latch onto the literal fox-and-grape imagery. Addressing this oversight, we present ``Mediom,'' a multilingual, multimodal idiom corpus of 3,533 Hindi, Bengali, and Thai idioms, each paired with gold-standard explanations, cross-lingual translations, and carefully aligned text–image representations. We benchmark both large language models (textual reasoning) and vision language models (figurative disambiguation) on Mediom, exposing systematic failures in metaphor comprehension. To mitigate these gaps, we propose ``HIDE,'' a Hinting-based Idiom Explanation framework that leverages error-feedback retrieval and targeted diagnostic cues for iterative reasoning refinement. Collectively, Mediom and HIDE establish a rigorous test bed and methodology for culturally grounded, multimodal idiom understanding embedded with reasoning hints in next-generation AI systems\footnote{Resources are available at \href{https://github.com/sarmistha-D/Hide-Mediom}{https://github.com/sarmistha-D/Hide}.}.
\end{abstract}


\section{Introduction}
\label{sec:intro}
\begin{figure}[t]    
    \centering
    \includegraphics[width=0.8\columnwidth]{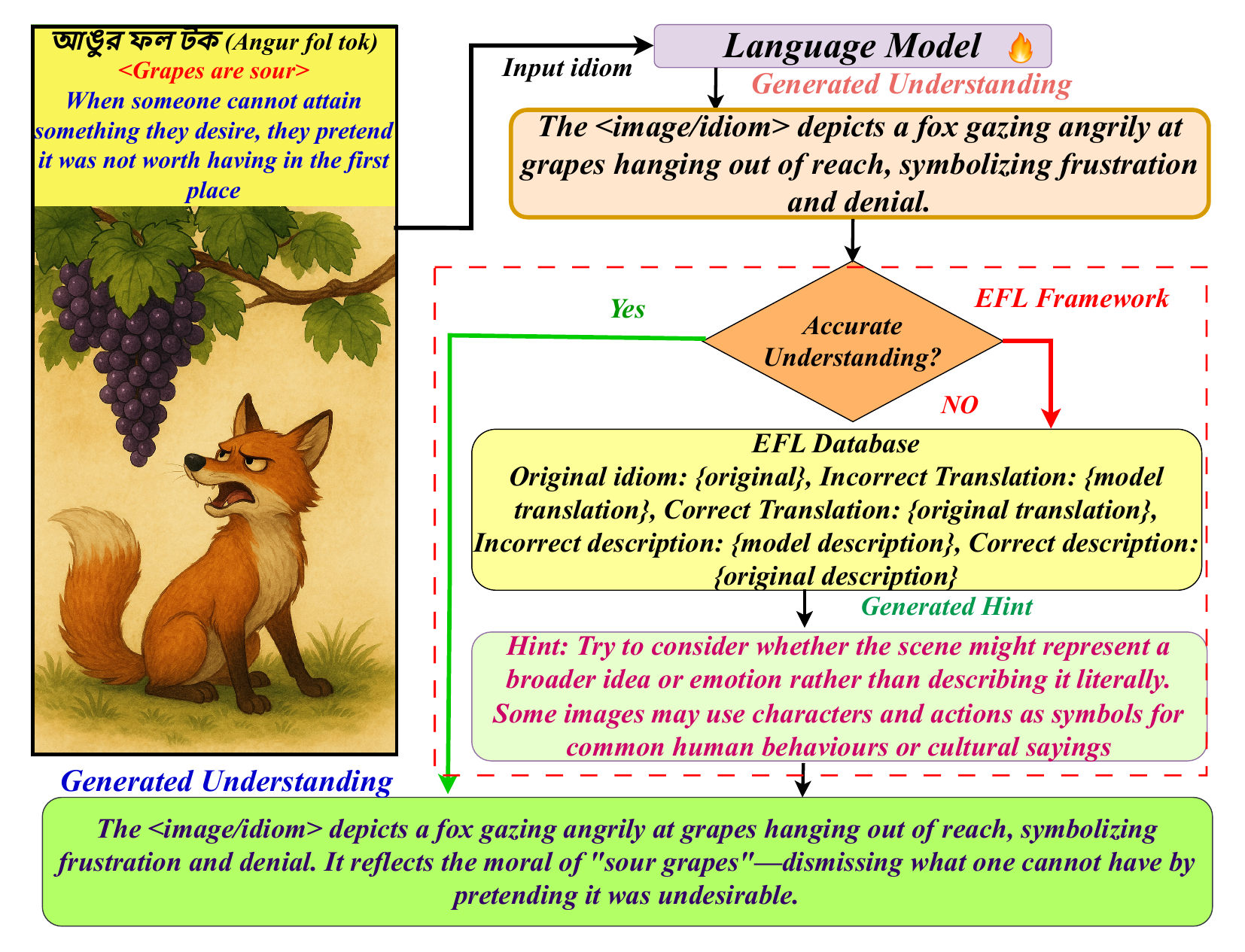}
    \caption{Idiomatic Understanding via HIDE inspired by EFL: The model first interprets the idiom, detects semantic errors, and uses targeted hints to refine and improve idiomatic comprehension.}
    \label{intro}
\end{figure}

Idioms embody figurative meanings that extend beyond their literal composition, conveying context-dependent semantics that are central to natural discourse. For instance, the Hindi idiom \begin{hindi}आसमान से गिरे, खजूर में अटके\end{hindi} (\textit{aasmaan se gire}, ``khajoor mein atke'')--literally, ``fallen from the sky, stuck in a date palm''--metaphorically denotes escaping one difficulty only to become trapped in another, often equally severe. Metaphorical competence is the human ability to fluidly traverse literal and figurative meanings; it underpins idiom comprehension. Yet as Honeck et al.~\cite{honeck2013proverb} demonstrate, such interpretations remain inherently context-dependent and subjective. Idioms, as crystallized metaphors, encapsulate cultural knowledge and linguistic economy. Recent advances demonstrate that Large Language Models (LLMs) achieve near-human performance across core linguistic tasks~\cite{kuribayashi2025large,NEURIPS2022_b1efde53}, while Vision Language Models (VLMs) further enhance reasoning by jointly modeling text and vision, yielding substantial gains in multimodal understanding such as visual question answering~\cite{lu2024wildvision}. 
However, idioms present a distinct challenge, as their meanings cannot be inferred from the literal definitions of individual words but instead depend on cultural context and conventional usage for proper understanding~\cite{fornaciari2024hard}. Idioms crystallize cultural knowledge, so accurate interpretation demands both linguistic and contextual fluency. For example, the Bengali idiom \begin{bengali}বারো মাসে তেরো পার্বণ\end{bengali} (\textit{baro mashe tero parbon}, ``Thirteen festivals in twelve months'') evokes a uniquely celebratory mindset, while the multilingual counterpart \textit{``walls have ears''} (Hindi \begin{hindi}दीवारों के भी कान होते हैं\end{hindi}, Bengali \begin{bengali}দেয়ালেরও কান আছে\end{bengali}, Thai \begin{thai}กำแพงมีหู ประตูมีตา\end{thai}) shows that shared metaphors transmit communal wisdom. Yet idiom research remains English-centric~\cite{haagsma2020magpie}, leaving Hindi, Bengali, Thai, and vision-grounded understanding largely unexplored. To fill this void, we release the first multimodal, cross-cultural idiom corpus, pairing expert-annotated text with contextual images for expressions. 
Additionally, we provide a hint-embedded idiom reasoning database that provides interpretive meanings, symbolic meanings, and context-specific cues. This supports a Hinting-based Idiom Explanation (HIDE) loop inspired by Error Feedback Learning (EFL)~\cite{zhang2024critic}, which helps correct generation errors during idiom interpretation as demonstrated in Figure~\ref{intro}. 

The research objectives of the current work are as follows:
\begin{enumerate*}[label=(\roman*) ]
    \item Evaluate the extent to which LLMs and VLMs encode culturally grounded knowledge and its impact on idiomatic inference accuracy.
    \item Analyze the contribution of the HIDE paradigm in enhancing fine-grained inferential reasoning over culturally nuanced idioms.
    \item Assess cross-model generalizability by benchmarking dataset performance across diverse LLM and VLM architectures.
\end{enumerate*}

Our contributions towards the research community are as follows:
\begin{enumerate*}[label=(\roman*) ]
    \item  We introduce ``Mediom,'' the first multimodal idiom benchmark for low-resource languages Hindi, Bengali, and Thai, comprising 3,533 idioms with fine-grained annotations and rigorously validated human explanations.
    \item  We propose ``HIDE,'' a hint-driven idiomatic explanation framework inspired by EFL.
    \item  We conduct two complementary evaluations: idiomatic understanding with LLMs and multimodal idiom interpretation with VLMs. 
\end{enumerate*}

\section{Background}
Recent natural language processing research has extensively explored figurative language comprehension and generation, including simile detection~\cite{zeng2020neural}, metaphor identification~\cite{chakrabarty2021mermaid}, pun recognition~\cite{poliak2018collecting}, and idiom retrieval~\cite{lee2016quote}.
Idiomatic expression modeling is closely related to idiom generation, contextual quotation recommendation, and literary text retrieval~\cite{wang2021continuity}. 
Subsequently, Qin et al.~\cite{qin2021ibert} developed a BERT-based model that encodes idiomatic expressions within both global and local contexts to improve the explanation of non-compositional meanings. Modern neural text generation models have significantly improved the contextual adaptation of idioms~\cite{adewumi2022vector}, with commonsense reasoning frameworks further enhancing figurative language understanding~\cite{bosselut2019comet}. HIDE advances this workflow by logging the model's past missteps and feeding them back as corrective context, iteratively tightening reasoning accuracy~\cite{zhang2024critic}. Despite growing interest in explanation strategies, their application to idiomatic understanding, particularly regarding the retention of cultural nuance and metaphorical coherence, remains largely unexplored.

To address the clear absence of idiom-centric resources for low-resource languages across both LLMs and VLMs exposed in Table~\ref{relatedwork}, our research introduces a retrieval-driven HIDE loop inspired by EFL that maintains a memory of past generation errors, retrieves structurally similar failures at inference time, and injects corrective semantic cues into subsequent prompts.

\begin{table}[t]
\caption{Comparative overview of the proposed Mediom against leading resources. Abbreviations: E = English; ML = multilingual set (Thai, Hindi, Bengali); Multimodality = paired text–image resources.}\label{relatedwork}
\centering
\scalebox{0.65}{
\begin{tabular}{lccccc}
\toprule
Corpus Name                                         & \multicolumn{1}{c}{Count} & Language & Explanations & Multimodality & Idiomatic Part \\ \midrule

SemEval-2013 \cite{korkontzelos2013semeval} & 4,350                                   & E           & $\times$    & $\times$  & $\times$               \\
FLUTE \cite{chakrabarty2022flute} & 8,962                                   & E           & $\checkmark$    & $\times$  & $\checkmark$(497)               \\
V-FLUTE\cite{saakyan2024v} & 6,027                                   & E & $\checkmark$ & $\checkmark$ & $\checkmark$(370)                \\
\textit{Mediom} (Proposed)                        & 3,533                                    & ML            & $\checkmark$                     & $\checkmark$    & 3,533              \\ \bottomrule
\end{tabular}}

\end{table}

\section{Corpus Formulation}\label{section:corpus}

\subsection{Data Collection}
Inspired by~\cite{haagsma2020magpie}, we initially compiled a dataset of 3,500 idioms from diverse linguistic and cultural backgrounds, sourcing them from online repositories, literature, and cultural archives, including Hindi idioms from the Simple Help\footnote{\href{https://thesimplehelp.com/hindi-idioms-with-meanings-and-sentences}{https://thesimplehelp.com/hindi-idioms-with-meanings}}, Bengali idioms from Bangla Probad~\footnote{\href{https://archive.org/details/in.ernet.dli.2015.455639/page/n557/mode/2up}{https://archive.org/details/in.ernet.dli.2015.455639}}, and Thai idioms~\cite{udomporn20145000}.
Our selection was designed to capture syntactic diversity across idioms in Hindi, Bengali, and Thai, which includes fixed expressions, such as \begin{thai}น้ำท่วมปาก\end{thai} (\textit{nam thuam pak}, ``unable to speak out,'') and verb-object structures, exemplified by \begin{hindi}नाक रगड़ना\end{hindi} (\textit{nāk ragaṛnā}), ``to plead intensely.'' We also included adjective-noun combinations, such as \begin{bengali}মিষ্টি স্বপ্ন\end{bengali} (\textit{miṣṭi swapna}, ``sweet dreams,'' prepositional phrases, such as \begin{thai}เข้าหูซ้ายทะลุหูขวา\end{thai} (\textit{khao hu sai thalu hu khwa}, ``in one ear and out the other,'' and binomial pairs, such as \begin{hindi}उल्टा सीधा\end{hindi} (\textit{ulṭā sīdhā}, ``all sorts of nonsense.'' For syntactically flexible idioms, normalization was performed by unifying verb inflections and replacing person-specific pronouns with neutral counterparts. However, structurally rigid idioms were preserved in their original form to maintain cultural and linguistic authenticity.

\begin{figure}[t]
    \centering
    \subfloat[\label{datasample}]{%
        \includegraphics[width=0.49\columnwidth]{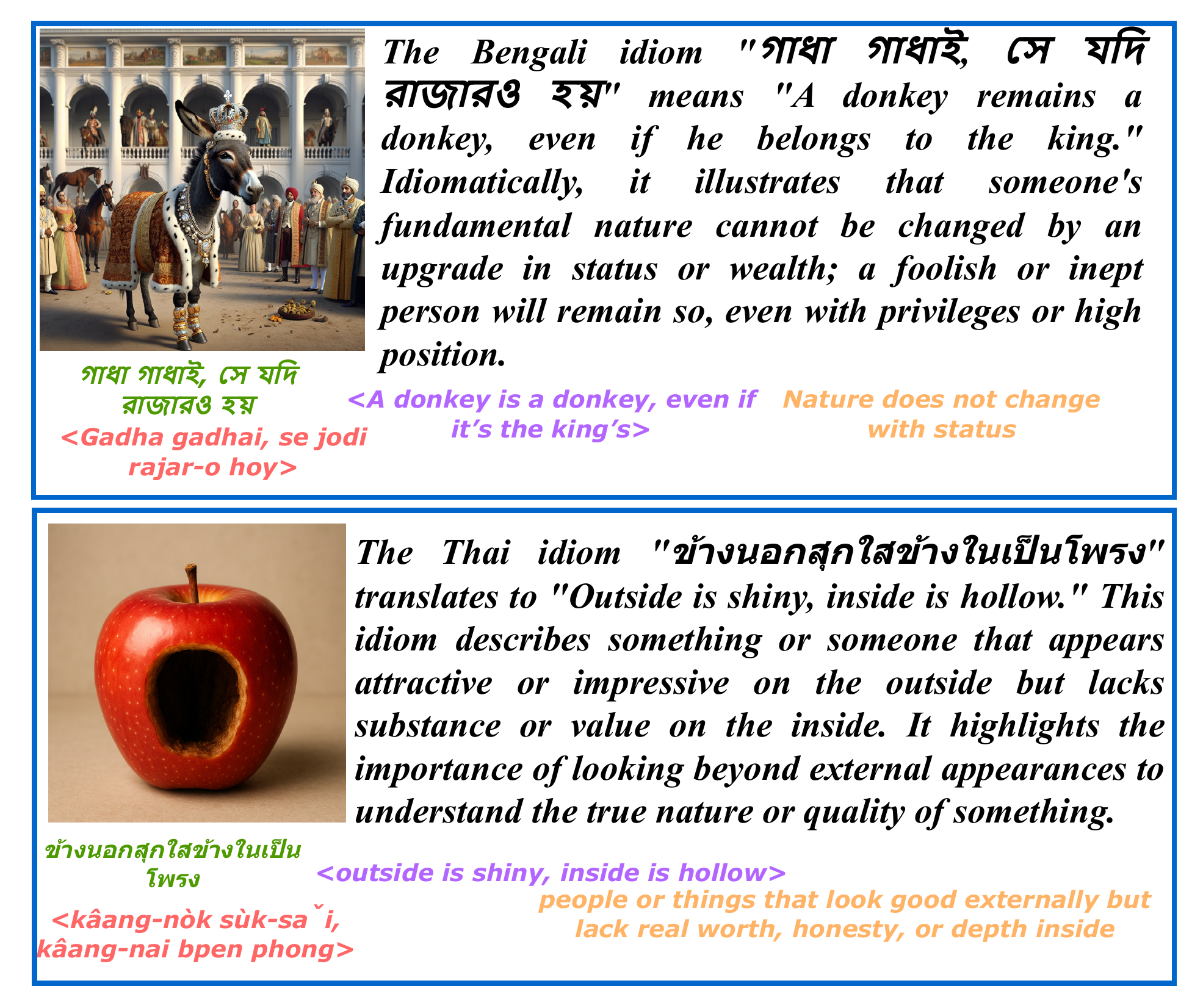}
    }%
    \hfill
    \subfloat[\label{expert}]{%
        \includegraphics[width=0.49\columnwidth]{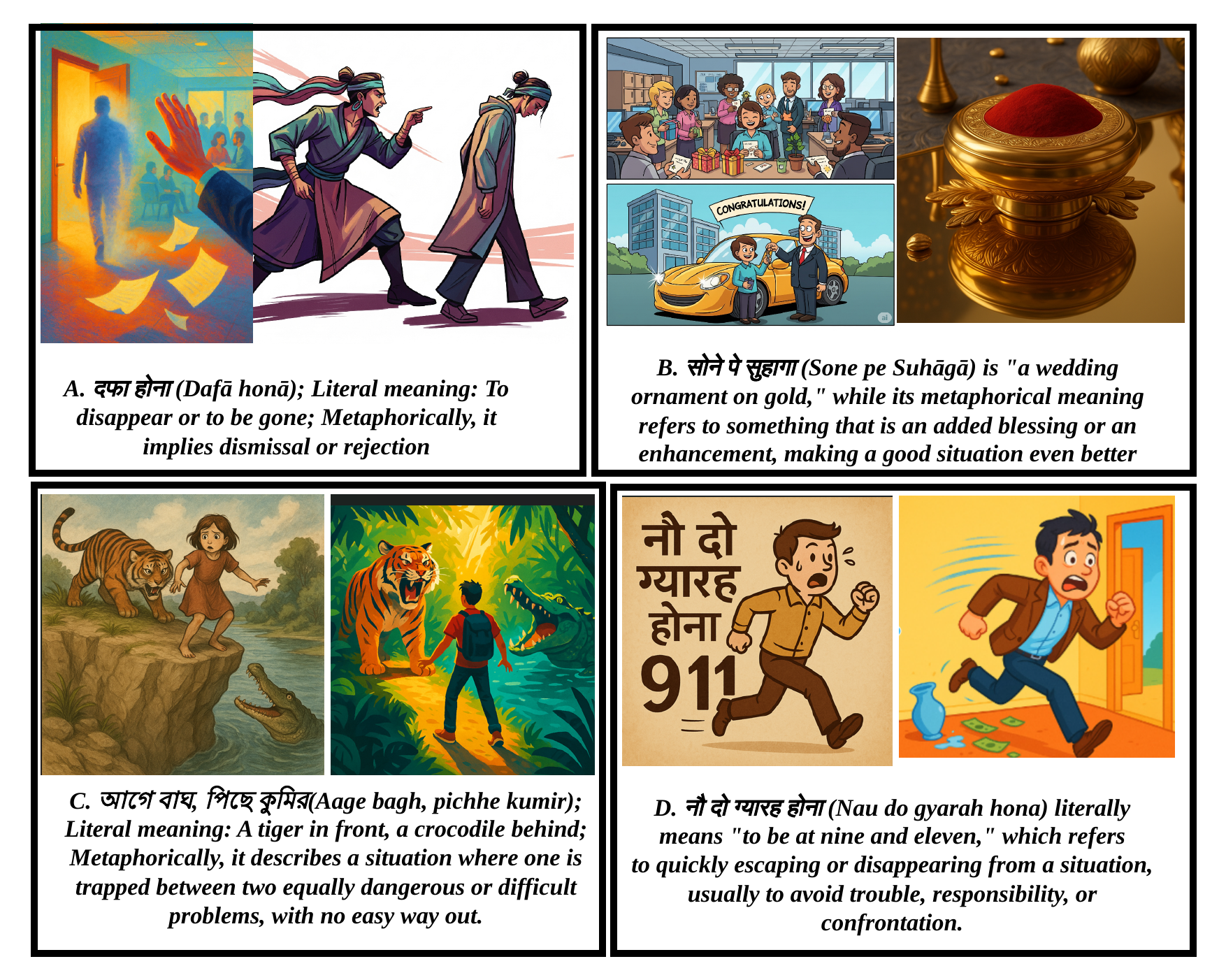}%
    }
    \caption{(a) Sample instances of our proposed Mediom dataset; (b) Two different candidate images for the same idiom in the Mediom dataset.}
    \label{fig:datasampleall}
\end{figure}

Following rigorous curation, 3,533 idioms were retained to ensure diversity and quality; representative samples are shown in Figure~\ref{datasample}.

\subsection{Data Quality Assurance}
To ensure rigorous annotation quality, we enlisted a doctoral scholar and two literature professors, native speakers of Thai, Hindi, and Bengali, whose combined academic and linguistic expertise delivers culturally nuanced, high-precision labels.
Initially, each expert annotated 100 samples, generating a reference set of 200 idiomatic explanations. Subsequently, we established Information Persistence Ratings (IPR) criteria to evaluate the quality of idiom annotations, which were structured around five core aspects:
\begin{enumerate*}[label=(\roman*)]
    \item Literal Translation, where each idiom was directly translated into English while preserving its original structure, metaphorical elements, and imagery; 
    \item Contextual Interpretation, in which annotators provided a succinct explanation of the idiom's meaning within its cultural and linguistic context;
    \item Usage Scenarios, which included brief examples illustrating real-world applications of the idiom for enhanced comprehension;
    \item Cultural Significance, where any historical, regional, or societal relevance associated with the idiom was documented; and 
    \item Coherence Preservation, ensuring the interpretation encapsulated the idiom's intended meaning without fragmentation. 
\end{enumerate*}
To ensure consistency and high annotation quality, linguistic experts established comprehensive guidelines following standard best practices. We then conducted a two-stage annotation competition with 15 participants: 
(i) a training phase using 100 reference samples, from which seven annotators qualified, and
(ii) a testing phase with an additional 100 samples, resulting in the selection of two final annotators. Annotators were compensated at \$0.5 per sample.

To further guarantee the reliability of the annotations, a stringent two-tiered validation protocol was implemented.
\begin{enumerate*}[label=(\roman*)]
    \item Peer Review: Each annotation was cross-verified by at least two additional annotators fluent in the source language to enhance accuracy and mitigate subjective biases. 
    \item Expert Validation: A panel comprising linguistic and cultural specialists possessing adequate linguistic knowledge on Thai, Hindi, and Bengali (can read and write proficiently) conducted a final review of the annotations. Their role included validating adherence to the IPR criteria for each sample. The rating is based on the retention of individual aspects. For instance, if five aspects are retained, the evaluation score will be five. 
    The annotation process yielded a robust inter-annotator agreement score of 0.82 (Cohen's kappa), reflecting substantial consistency and reliability across the dataset. 
\end{enumerate*}

\subsection{Idiomatic Image Creation}
We implement a visual-idiom generation module. Since no curated visual idiom corpus exists, we synthesize high-fidelity images with a prompt-driven pipeline: gold-standard idiom explanations feed DALL·E 3~\cite {openai2023dalle3} via GPT-4o prompts~\cite {hurst2024gpt}. Each prompt is crafted to preserve both figurative meaning and cultural nuance. Figure~\ref{expert} illustrates the refinement process:
\begin{enumerate*}[label=(\roman*)]
    \item for the Hindi idiom \begin{hindi}सोने पे सुहागा\end{hindi} (``icing on the cake''), literal visuals were replaced with metaphorically aligned depictions; 
    \item for the Bengali idiom \begin{bengali}আগে বাঘ, পিছে কুমির\end{bengali}, only frames conveying an inescapable dilemma were retained; and
    \item for the Hindi idiom \begin{hindi}नौ दो ग्यारह होना\end{hindi}, arithmetic literalizations (e.g., ``9+2=11'') were discarded.
\end{enumerate*}
In such cases, annotators refined prompts with contextual cues and regenerated images to ensure cultural fidelity and semantic alignment.
    
\section{Methodology}
\begin{figure*}[t]    
    \centering
    \includegraphics[width=0.70\textwidth]{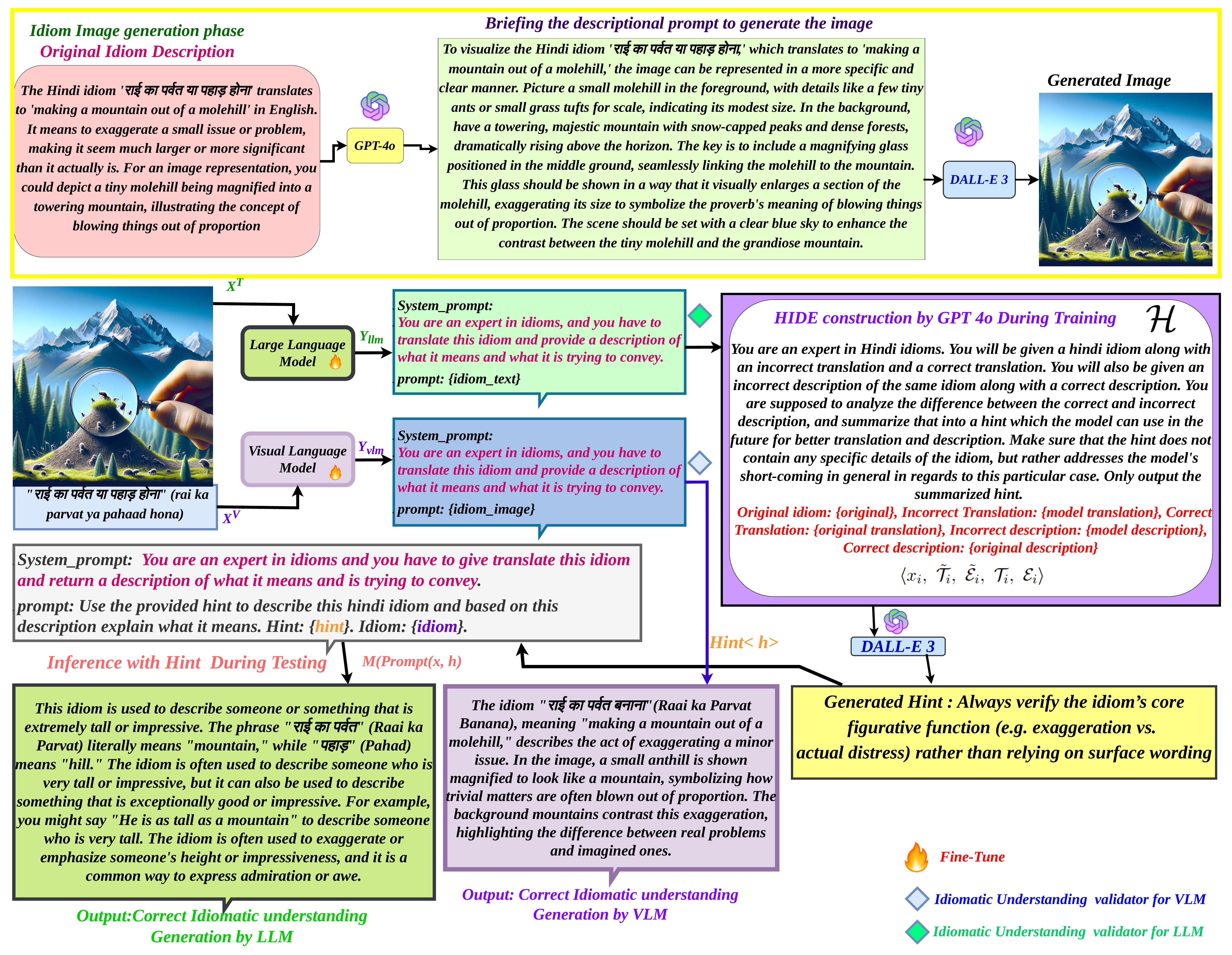}
    \caption{Architectural framework for idiom explanation that fuses LLMs and VLMs, augmented by a HIDE module with Hint Generation}
    \label{archi_HIDE}
\end{figure*}

Given an idiom pair $(x_t, x_i)$, where $x_t$ is the textual representation and $x_i$ is the corresponding image, this study aims to develop a multimodal idiom interpretation framework leveraging LLMs and VLMs. The LLM-based model $P_{\theta}(y_t \mid x_t)$ is trained via supervised learning to generate textual idiom interpretations, while the VLM-based model $P_{\phi}(y_t \mid x_i)$ learns idiom meanings from images using supervised learning without preference optimization. The dataset consists of text-meaning pairs $\mathcal{D}_t = \{(x_{t_j}, y_j)\}_{j=1}^{N}$ and image-meaning pairs $\mathcal{D}_i = \{(x_{i_j}, y_j)\}_{j=1}^{N}$ where $N$ is the number of idioms. To achieve our objective, the first phase centers on idiom explanation, leveraging both LLMs and VLMs to capture nuanced semantic and visual cues. The second phase employs the hinting database embedded in the HIDE framework to iteratively refine model performance and enhance interpretability. Figure~\ref{archi_HIDE} describes the entire methodological representation.

\subsection{Idiom Explanation Generation}
We fine-tuned the Gemma2-9B~\cite{gemma_2024} using supervised learning on idiom--explanation pairs $(\mathcal{X}^{(\tau)}, \mathcal{Y})$, where $\mathcal{X}^{(\tau)}$ represents a figurative idiom in textual form and $\mathcal{Y}$ denotes a corresponding human-curated explanation. Following fine-tuning, Gemma2-9B underwent human evaluation to assess outputs for semantic precision and sociocultural alignment.

In parallel, to enable visual idiom interpretation, we fine-tuned a pre-trained VLM, Paligemma2-10B~\cite{steiner2024paligemma}, denoted as $P_{\phi}(\mathcal{Y} \mid \mathcal{X}^{(\nu)})$, where $\mathcal{X}^{(\nu)}$ is a visual input associated with an idiom (e.g., metaphorical illustrations or idiom-linked scenes), and $\mathcal{Y}$ is the textual explanation predicted by the model. However, post-fine-tuning analysis revealed multiple instances of generative misinterpretation. For instance, the Bengali idiom \begin{bengali}
মিষ্টি মুখে জুতো মারা\end{bengali} (\textit{``mishti mukhe jutor bari}'')-literally, ``Hitting with a shoe while smiling sweetly''--was misinterpreted by the system as commentary on conflicting emotional expressions or unusual behavior involving footwear. This interpretation failed to capture the idiom's actual implication--the act of harming or insulting someone with deceptive politeness or a pretence of affection. Among a held-out evaluation set of 600 idiomatic samples, 360 instances (60\%) exhibited comparable inaccuracies, often resulting from insufficient grounding in sociocultural nuance or an over-literal mapping from visual or lexical features. A granular inspection of the models' outputs revealed that their inferential reasoning trajectories frequently diverge from the gold-standard at the very onset of generation. This early misalignment necessitates the introduction of controlled nudges to systematically steer the reasoning away from known pitfalls.

To address this, we integrated a HIDE mechanism. In this framework, idiomatic explanations generated by the fine-tuned LLM ($\mathcal{Y}_{\text{LLM}}$) are passed into the HIDE framework. This verifier-guided refinement loop enforces canonical idiom alignment and cultural fidelity, substantially boosting the semantic and contextual accuracy of the language models.

\subsection{HIDE construction inspired by EFL}
\noindent To construct hint-embedded idiomatic explanations, initially after inference, each incorrectly handled idiom is ingested as the quintuple $\langle x_{i},\;\tilde{\mathcal{T}}_{i},\;\tilde{\mathcal{E}}_{i},\;\mathcal{T}_{i},\;\mathcal{E}_{i}\rangle$, where $x_{i}$ is the idiom string, $(\tilde{\mathcal{T}}_{i},\tilde{\mathcal{E}}_{i})$ are the model-generated translation and explanation, and $(\mathcal{T}_{i},\mathcal{E}_{i})$ are their human-annotated gold counterparts. A discriminator compares $\tilde{\mathcal{E}}_{i}$ with $\mathcal{E}_{i}$ and compresses the discrepancy into a high-level corrective hint $h_{i}=\phi\!\bigl(\tilde{\mathcal{E}}_{i},\mathcal{E}_{i}\bigr)$. The idiom is then embedded by a semantic encoder $f\!:\!\mathcal{X}\!\to\!\mathbb{R}^{d}$, producing $z_{i}=f(x_{i})$ where $d$ is the embedding dimensional. The tuple $\langle z_{i},\,h_{i},\,\tilde{\mathcal{T}}_{i},\,\tilde{\mathcal{E}}_{i}\rangle$ is archived in an \emph{Idiomatic Error-Feedback Repository} (denoted $\mathcal{H}$) for future reuse.

\noindent At the secondary inference time, the system treats a test idiom $x$ as a query: it computes $z=f(x)$ and retrieves the repository entry $j=\arg\max_{k}\cos\!\bigl(z,z_{k}\bigr)$ that is most similar in embedding space. The accompanying hint $h_{j}$ is concatenated with $x$ to form an augmented prompt, $\text{Prompt}(x,h_{j}) = x \,\Vert\, h_{j}$, which is passed to the generation model $M$ to yield the final translation–explanation pair $\hat{y}=(\hat{\mathcal{T}},\hat{\mathcal{E}})=M\!\bigl(\text{Prompt}(x,h_{j})\bigr)$. Injecting this retrieved-error context exposes the model to a concrete, previously mis-handled scenario that is semantically close to the current input, steering its reasoning away from known pitfalls while preserving the original flow.

\section{Experimental Results and Discussion}
This section details the experimental protocol, baseline settings, and a direct comparative analysis of LLMs and VLMs, complemented by qualitative error diagnostics that expose interpretive limitations and open research challenges. Our study addresses three focal Research Questions (RQs):
RQ1: Idiomatic Competence. How effectively do LLMs and VLMs internalize and interpret culturally grounded, metaphor-rich expressions?
RQ2: HIDE+EFL Impact. How effectively does HIDE enhance idiom understanding and contextual reasoning in LLMs without degrading overall performance?
RQ3: Generalizability and Implications. How robustly does the dataset generalize across model families, and what broader socio-linguistic insights does it reveal?

Models were fine-tuned for 5 epochs with a learning rate of \(1\times10^{-5}\) on an NVIDIA A100 (80\,GB) GPU, using the AdamW optimizer. Training employed a batch size of 8 with two-step gradient accumulation and native Automatic Mixed Precision (AMP) for memory-efficient mixed-precision execution. A linear learning-rate scheduler was applied to ensure stable convergence, while Top-\(K\) sampling (\(K{=}10\)) encouraged generative diversity. The dataset was split into 80\% training and 20\% testing. The model achieved a final training cross-entropy loss of 2.1986. A comparative evaluation was performed across a range of prominent LLMs and VLMs, including GPT-3.5~\cite{brown2020language}, Mistral-7B~\cite{jiang2023mistral}, LLaMA2-7B~\cite{touvron2023llama}, Blip2-7B~\cite{li2023blip2}, Qwen2-VL-7B-Instruct~\cite{Qwen2VL}, SmolVLM-Instruct~\cite{huggingface2025smolvlm500m}, Video-LLaVA-7B~\cite{zhu2023languagebind}.
Model outputs were benchmarked using an extensive suite of evaluation metrics that captured lexical overlap, semantic alignment, and distributional similarity as reported in Table~\ref{comp}.

\begin{table*}[t]
\caption{Performance variance across baseline LLMs and VLM configurations. R-1, R-2, and R-L denote ROUGE-1, ROUGE-2, and ROUGE-L; B-1, B-2, B-3, and B-L denote BLEU-1, BLEU-2, BLEU-3, and BLEU-L; BS denotes BERTScore. MS, CD, JSD, L2, L1, PS, and FRS represent METEOR, Cosine Distance, Jensen--Shannon Divergence, L2 (Euclidean) Distance, L1 (Manhattan) Distance, Perplexity Score, and Flesch--Kincaid Readability Score. Boldface indicates the best results; $\uparrow$ higher is better, $\downarrow$ lower is better.}
\label{comp}
\centering
\resizebox{0.92\textwidth}{!}{%
\begin{tabular}{@{}clrrrrrrrrrrrrrrr@{}} 
\toprule 
\multirow{2}{*}{\begin{tabular}[c]{@{}c@{}}Model Type\end{tabular}} & 
\multirow{2}{*}{Model} & 
\multicolumn{15}{c}{Metrics} \\ 
\cmidrule(l){3-17} 
& & 
\multicolumn{1}{c}{R-1} & 
\multicolumn{1}{c}{R-2} & 
\multicolumn{1}{c}{R-L} & 
\multicolumn{1}{c}{B-1} & 
\multicolumn{1}{c}{B-2} & 
\multicolumn{1}{c}{B-3} & 
\multicolumn{1}{c}{B-L} & 
\multicolumn{1}{c}{BS} & 
\multicolumn{1}{c}{MS} & 
\multicolumn{1}{c}{CD} & 
\multicolumn{1}{c}{JSD} & 
\multicolumn{1}{c}{L2} & 
\multicolumn{1}{c}{L1} & 
\multicolumn{1}{c}{PS} & 
\multicolumn{1}{c}{FRS} \\ 
\midrule 

LLM & LLaMA2-7B & 0.42 & 0.18 & 0.34 & 0.24 & 0.16 & 0.13 & 0.21 & 0.67 & 0.37 & 0.39 & 0.58 & 12.17 & 98.95 & 74.37 & 58.11 \\ 
& Mistral -7B & 0.47 & 0.24 & 0.41 & 0.27 & 0.24 & 0.20 & 0.27 & 0.73 & 0.45 & 0.34 & 0.53 & 9.82 & 78.91 & 70.11 & 38.05 \\ 
& GPT 3.5 & 0.51 & 0.25 & 0.42 & 0.31 & 0.24 & 0.20 & 0.27 & 0.75 & 0.46 & 0.31 & 0.54 & 10.14 & 49.73 & 68.31 & 48.91 \\ 
& Gemma2-9B & 0.65 & 0.55 & 0.52 & 0.51 & 0.46 & 0.44 & 0.55 & 0.79 & 0.55 & 0.22 & 0.39 & 7.19 & 44.13 & 57.69 & 67.15 \\ 
& LLaMA2-7B+HIDE & 0.42 & 0.18 & 0.35 & 0.24 & 0.18 & 0.14 & 0.21 & 0.69 & 0.38 & 0.38 & 0.57 & 11.93 & 96.02 & 73.23 & 59.91 \\ 
& Mistral -7B+HIDE & 0.48 & 0.24 & 0.42 & 0.31 & 0.25 & 0.21 & 0.29 & 0.74 & 0.46 & 0.32 & 0.53 & 9.76 & 76.56 & 68.71 & 39.23 \\ 
& Gemma2-9B+HIDE & \textbf{0.68}$\uparrow$ & \textbf{0.56}$\uparrow$ & \textbf{0.58}$\uparrow$ & \textbf{0.53}$\uparrow$ & \textbf{0.50}$\uparrow$ & \textbf{0.48}$\uparrow$ & \textbf{0.57}$\uparrow$ & \textbf{0.81}$\uparrow$ & \textbf{0.56}$\uparrow$ & \textbf{0.21}$\downarrow$ & \textbf{0.38}$\downarrow$ & \textbf{7.01}$\downarrow$ & \textbf{43.71}$\downarrow$ & \textbf{56.81}$\downarrow$ & \textbf{69.23}$\uparrow$ \\ 
\midrule 

VLM & Blip2-7B & 0.15 & 0.03 & 0.12 & 0.18 & 0.08 & 0.06 & 0.02 & 0.47 & 0.05 & 0.58 & 0.70 & 20.34 & \textbf{148.98}$\downarrow$ & 175.83 & \textbf{74.54}$\uparrow$ \\ 
& Qwen2-VL-7B-Instruct & 0.36 & 0.09 & 0.22 & 0.36 & 0.18 & 0.11 & 0.06 & 0.60 & 0.21 & 0.28 & 0.60 & 18.08 & 185.43 & 109.27 & 51.46 \\ 
& SmolVLM-Instruct & 0.37 & 0.11 & 0.22 & 0.33 & 0.18 & 0.11 & 0.06 & 0.62 & 0.25 & 0.25 & 0.57 & 29.47 & 249.61 & 114.41 & 51.82 \\ 
& Video-LLaVA-7B & 0.38 & 0.10 & 0.24 & 0.37 & 0.19 & 0.11 & 0.06 & 0.62 & 0.22 & 0.32 & 0.59 & 18.20 & 175.66 & 131.72 & 50.64 \\ 
& Paligemma2-10B & \textbf{0.43}$\uparrow$ & \textbf{0.14}$\uparrow$ & \textbf{0.27}$\uparrow$ & \textbf{0.41}$\uparrow$ & \textbf{0.24} & 0.15 & 0.09 & \textbf{0.66}$\uparrow$ & \textbf{0.28}$\uparrow$ & \textbf{0.24}$\downarrow$ & \textbf{0.56}$\downarrow$ & 18.97 & 185.90 & 86.09 & 52.30 \\ 
\bottomrule 
\end{tabular}}
\end{table*}

\subsection{Resultant Discussion}
This section synthesizes the findings in response to the stated RQs, supported by qualitative insights and error analyzes across model generations.

\subsubsection{Response to RQ1: Idiomatic competence across LLMs and VLMs}%
Table~\ref{comp} highlights a clear performance divide between text-only LLMs and VLMs on culturally dense idioms. Even without error-driven refinement, Gemma2-9B leads across all metrics, substantially outperforming the strongest VLM, Paligemma2-10B. Distance-based measures further confirm this gap, with lower cosine distance and Jensen–Shannon divergence for LLMs, indicating tighter semantic alignment with figurative meanings. Superior readability and lower perplexity likewise favor LLMs, underscoring that large-scale textual pretraining remains more effective than current multimodal grounding for internalizing culturally encoded idiomatic semantics in Hindi, Bengali, and Thai.

\subsubsection{Response to RQ2: HIDE+EFL gains in idiom comprehension}%
Injecting Error-Feedback Learning (via the HIDE retriever) converts past errors into micro-lessons that significantly sharpen idiom handling. For LLMs, Gemma2-9B + HIDE raises ROUGE-1/2/3, trims cosine distance to 0.21, and cuts L2/L1 errors while lifting readability (FRS ≈ 69) and lowering perplexity. Mistral-7B and LLaMA2-7B record smaller yet consistent gains, accompanied by reduced perplexity, proving that even lightweight hints nudge mid-sized models toward the correct figurative space.

\subsubsection{Response to RQ3: Dataset Impact}
The Mediom corpus provides a multimodal benchmark for idiomatic comprehension in Hindi, Bengali, and Thai by coupling high-resolution images with expert-curated explanations. This visual grounding sharpens cross-lingual transfer (e.g., idiom translation) and equips conversational agents with culturally aligned reasoning. In education, image-anchored idioms turn abstract metaphors into concrete, engaging content, boosting learner retention and motivation.

\subsection{Analytical Discussion}
\subsubsection{Human Evaluation}
Three native linguists evaluated 300 idiom instances across five dimensions: literal accuracy, contextual fit, usage naturalness, cultural depth, and overall coherence. Fine-tuned LLMs (e.g., GPT-3.5, Gemma2-9B) and VLMs (e.g., Paligemma2-10B, Qwen2-VL-7B-Instruct) excel in \textit{literal translation} and \textit{contextual interpretation}, reflecting strong linguistic priors, but initially lag in \textit{cultural significance} and \textit{coherence}. Incorporating HIDE yields substantial gains, particularly for LLMs such as Gemma2-9B and LLaMA2-7B, by converting prior reasoning errors into structured semantic cues, thereby improving metaphor comprehension, cultural alignment, and usage coherence. While HIDE-enabled LLMs achieve the most consistent idiomatic reasoning, only fine-tuned VLMs exhibit marked improvements in visual–semantic alignment and contextual depth, narrowing the unimodal–multimodal performance gap.

\begin{figure}[t]
    \centering
    \subfloat[Qualitative analysis of the Bengali idiom
        \begin{bengali}আগে বাঘ, পিছে কুমির\end{bengali}
        (\textit{Aage bagh, pichhe kumir}).
        Literal: A tiger in front, a crocodile behind.
        Metaphorical: Trapped between two equally dangerous situations.%
        \label{fig:qualitative_bengali}]{%
        \includegraphics[width=0.48\columnwidth]{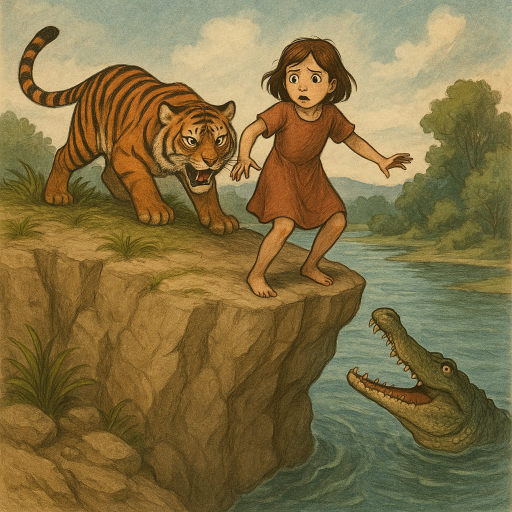}%
    }%
    \hfill
    \subfloat[Error analysis of the Hindi idiom
        \begin{hindi}नाच न जाने आँगन टेढ़ा\end{hindi}
        (\textit{Naach na jaane angan tedha}).
        Literal: One who cannot dance blames the courtyard.
        Metaphorical: Blaming external factors for personal failure.%
        \label{img:error}]{%
        \includegraphics[width=0.47\columnwidth]{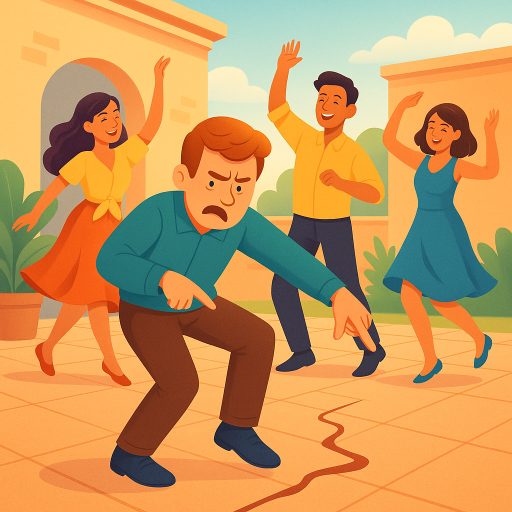}%
    }

    \caption{Head-to-head qualitative and error analysis.}
    \label{fig:idiom_analysis}
\end{figure}

\subsubsection{Qualitative Analysis}
Figure~\ref{fig:qualitative_bengali} illustrates qualitative differences in Bengali idiom interpretation. Gemma2-9B augmented with HIDE delivers the most nuanced and context-aware explanation, effectively capturing the idiom's core dilemma of simultaneous threats. Mistral-7B and LLaMA2-7B provide competent but less granular interpretations. In contrast, multimodal models such as VideoLLaVA-7B enhance comprehension through vivid visualizations of danger and tension, while BLIP-2, Qwen2-VL-7B-Instruct, Paligemma2-10B, and SmolVLM-Instruct reinforce the concept visually but fall short in figurative depth. Overall, HIDE-equipped Gemma2-9B best captures idiomatic semantics, whereas VLMs primarily excel in visual grounding.

\subsubsection{Error Analysis}
Figure~\ref{img:error} underscores a clear modality gap. While LLMs fine-tuned with HIDE (Gemma2-9B, Mistral-7B, LLaMA2-7B) reliably captured the core metaphor (blaming the courtyard for one’s own lack of skill), framing it within themes of accountability and self-reflection. GPT-3.5, however, slipped into over-literal narration, anchoring on irrelevant spatial details and losing the abstraction. Conclusively, HIDE equips LLMs to internalize cultural metaphor, whereas current VLMs struggle to transcend literal scene parsing.

\section{Conclusion}
In this work, we introduce Mediom, a multimodal dataset containing 3,533 idioms across Hindi, Thai, and Bengali, paired with gold-standard human explanations and aligned visual representations. Our analysis reveals that HIDE, inspired by EFL fine-tuning, slightly enhances lexical and contextual reasoning in LLMs (LLaMA2-7B, Gemma2-9B, Mistral-7B), facilitating nuanced idiomatic interpretations. Moreover, VLMs (Paligemma2-10B) effectively leverage visual cues to capture cultural subtleties. Mediom thus establishes a benchmark for evaluating culturally and figuratively aware AI models, advancing idiomatic comprehension through an integrated textual, visual, and cultural framework.

\bibliographystyle{IEEEbib}
\bibliography{references_short}

\begin{thebibliography}{10}

\bibitem{honeck2013proverb}
Richard~P. Honeck,
\newblock {\em A Proverb in Mind: The Cognitive Science of Proverbial Wit and Wisdom},
\newblock Lawrence Erlbaum Associates, Mahwah, NJ, USA, 1997.

\bibitem{kuribayashi2025large}
Tatsuki Kuribayashi, Yohei Oseki, Souhaib~Ben Taieb, Kentaro Inui, and Timothy Baldwin,
\newblock ``Large language models are human-like internally,''
\newblock {\em Trans. Assoc. Comput. Linguist.}, vol. 13, pp. 1743--1766, 2025.

\bibitem{NEURIPS2022_b1efde53}
Long Ouyang, Jeff Wu, Xu~Jiang, Diogo Almeida, Carroll~L. Wainwright, Pamela Mishkin, Chong Zhang, et~al.,
\newblock ``Training language models to follow instructions with human feedback,''
\newblock in {\em Advances in Neural Information Processing Systems}, 2022, vol.~35, pp. 27730--27744.

\bibitem{lu2024wildvision}
Yujie Lu, Dongfu Jiang, Wenhu Chen, William~Yang Wang, Yejin Choi, and Bill~Yuchen Lin,
\newblock ``{WildVision}: Evaluating vision-language models in the wild with human preferences,''
\newblock in {\em Advances in Neural Information Processing Systems}, 2024, vol.~38.

\bibitem{fornaciari2024hard}
Francesca De~Luca Fornaciari, Bego{\~n}a Altuna, Itziar Gonzalez-Dios, and Maite Melero,
\newblock ``A hard nut to crack: Idiom detection with conversational large language models,''
\newblock in {\em The Workshop on Figurative Language Processing}, 2024, pp. 35--44.

\bibitem{haagsma2020magpie}
Hessel Haagsma, Johan Bos, and Malvina Nissim,
\newblock ``{MAGPIE}: A large corpus of potentially idiomatic expressions,''
\newblock in {\em The Language Resources and Evaluation Conference}, 2020, pp. 279--287.

\bibitem{zhang2024critic}
Di~Zhang, Junxian Li, Jingdi Lei, Xunzhi Wang, Yujie Liu, Zonglin Yang, Jiatong Li, et~al.,
\newblock ``{Critic-V}: {VLM} critics help catch {VLM} errors in multimodal reasoning,''
\newblock in {\em The IEEE/CVF Conference on Computer Vision and Pattern Recognition}, 2025.

\bibitem{zeng2020neural}
Jiali Zeng, Linfeng Song, Jinsong Su, Jun Xie, Wei Song, and Jiebo Luo,
\newblock ``Neural simile recognition with cyclic multitask learning and local attention,''
\newblock in {\em The AAAI Conference on Artificial Intelligence}, 2020, pp. 9515--9522.

\bibitem{chakrabarty2021mermaid}
Tuhin Chakrabarty, Xurui Zhang, Smaranda Muresan, and Nanyun Peng,
\newblock ``{MERMAID}: Metaphor generation with symbolism and discriminative decoding,''
\newblock in {\em The Conference of the North American Chapter of the ACL}, 2021, pp. 4250--4261.

\bibitem{poliak2018collecting}
Adam Poliak, Aparajita Haldar, Rachel Rudinger, J.~Edward Hu, Ellie Pavlick, Aaron~Steven White, and Benjamin~Van Durme,
\newblock ``Collecting diverse natural language inference problems for sentence representation evaluation,''
\newblock in {\em The Conference on Empirical Methods in Natural Language Processing}, 2018, pp. 67--81.

\bibitem{lee2016quote}
Hanbit Lee, Yeonchan Ahn, Haejun Lee, Seungdo Ha, and Sang goo Lee,
\newblock ``Quote recommendation in dialogue using deep neural network,''
\newblock in {\em The International ACM SIGIR Conference on Research and Development in Information Retrieval}, 2016, pp. 957--960.

\bibitem{wang2021continuity}
Lingzhi Wang, Jing Li, Xingshan Zeng, Haisong Zhang, and Kam-Fai Wong,
\newblock ``Continuity of topic, interaction, and query: Learning to quote in online conversations,''
\newblock in {\em The Conference on Empirical Methods in Natural Language Processing}, 2020, pp. 6640--6650.

\bibitem{qin2021ibert}
Ruiyang Qin, Haozheng Luo, Zheheng Fan, and Ziang Ren,
\newblock ``{IBERT}: Idiom cloze-style reading comprehension with attention,''
\newblock {\em arXiv preprint arXiv:2112.02994}, 2021.

\bibitem{adewumi2022vector}
Tosin Adewumi, Foteini Liwicki, and Marcus Liwicki,
\newblock ``Vector representations of idioms in conversational systems,''
\newblock {\em Sci}, vol. 4, no. 4, pp. 37, 2022.

\bibitem{bosselut2019comet}
Antoine Bosselut, Hannah Rashkin, Maarten Sap, Chaitanya Malaviya, Asli Celikyilmaz, and Yejin Choi,
\newblock ``{COMET}: Commonsense transformers for automatic knowledge graph construction,''
\newblock in {\em The Annual Meeting of the ACL}, 2019, pp. 4762--4779.

\bibitem{korkontzelos2013semeval}
Ioannis Korkontzelos, Torsten Zesch, Fabio~Massimo Zanzotto, and Chris Biemann,
\newblock ``{SemEval}-2013 task 5: Evaluating phrasal semantics,''
\newblock in {\em The International Workshop on Semantic Evaluation}, 2013, pp. 39--47.

\bibitem{chakrabarty2022flute}
Tuhin Chakrabarty, Arkadiy Saakyan, Debanjan Ghosh, and Smaranda Muresan,
\newblock ``{FLUTE}: Figurative language understanding through textual explanations,''
\newblock in {\em The Conference on Empirical Methods in Natural Language Processing}, 2022, pp. 7139--7159.

\bibitem{saakyan2024v}
Arkadiy Saakyan, Shreyas Kulkarni, Tuhin Chakrabarty, and Smaranda Muresan,
\newblock ``Understanding figurative meaning through explainable visual entailment,''
\newblock in {\em The Conference of the Nations of the Americas Chapter of the ACL}, 2025, pp. 1--23.

\bibitem{udomporn20145000}
Ekarat Udomporn,
\newblock {\em 5000 Thai Idioms: From the Past Right on up to Now!},
\newblock P.S. Pattana Publishing, 2014.

\bibitem{openai2023dalle3}
{OpenAI},
\newblock ``Improving image generation with better captions,''
\newblock Tech. {R}ep., OpenAI, 2023.

\bibitem{hurst2024gpt}
Aaron Hurst, Adam Lerer, Adam~P. Goucher, Adam Perelman, Aditya Ramesh, Aidan Clark, AJ~Ostrow, et~al.,
\newblock ``{GPT-4o} system card,''
\newblock {\em arXiv preprint arXiv:2410.21276}, 2024.

\bibitem{gemma_2024}
{Gemma Team},
\newblock ``Gemma,''
\newblock {\em Kaggle}, 2024.

\bibitem{steiner2024paligemma}
Andreas Steiner, Andr{\'e}~Susano Pinto, Michael Tschannen, Daniel Keysers, Xiao Wang, Yonatan Bitton, et~al.,
\newblock ``{PaliGemma} 2: A family of versatile {VLMs} for transfer,''
\newblock {\em arXiv preprint arXiv:2412.03555}, 2024.

\bibitem{brown2020language}
Tom~B. Brown, Benjamin Mann, Nick Ryder, Melanie Subbiah, Jared Kaplan, Prafulla Dhariwal, Arvind Neelakantan, et~al.,
\newblock ``Language models are few-shot learners,''
\newblock in {\em Advances in Neural Information Processing Systems}, 2020, vol.~33, pp. 1877--1901.

\bibitem{jiang2023mistral}
Albert~Q. Jiang, Alexandre Sablayrolles, Arthur Mensch, Chris Bamford, Devendra~Singh Chaplot, Diego de~las Casas, Florian Bressand, et~al.,
\newblock ``Mistral 7{B},''
\newblock {\em arXiv preprint arXiv:2310.06825}, 2023.

\bibitem{touvron2023llama}
Hugo Touvron, Louis Martin, Kevin Stone, Peter Albert, Amjad Almahairi, Yasmine Babaei, Nikolay Bashlykov, et~al.,
\newblock ``Llama 2: Open foundation and fine-tuned chat models,''
\newblock {\em arXiv preprint arXiv:2307.09288}, 2023.

\bibitem{li2023blip2}
Junnan Li, Dongxu Li, Silvio Savarese, and Steven Hoi,
\newblock ``{BLIP-2}: Bootstrapping language-image pre-training with frozen image encoders and large language models,''
\newblock in {\em The International Conference on Machine Learning}, 2023, vol. 202, pp. 19730--19742.

\bibitem{Qwen2VL}
Peng Wang, Shuai Bai, Sinan Tan, Shijie Wang, Zhihao Fan, Jinze Bai, Keqin Chen, Xuejing Liu, Jialin Wang, et~al.,
\newblock ``{Qwen2-VL}: Enhancing vision-language model's perception of the world at any resolution,''
\newblock {\em arXiv preprint arXiv:2409.12191}, 2024.

\bibitem{huggingface2025smolvlm500m}
{Hugging Face},
\newblock ``{SmolVLM-500M-Base},'' \url{https://huggingface.co/HuggingFaceTB/SmolVLM-500M-Base}, 2025.

\bibitem{zhu2023languagebind}
Bin Zhu, Bin Lin, Munan Ning, Yang Yan, Jiaxi Cui, Hongfa Wang, Yatian Pang, Wenhao Jiang, Junwu Zhang, et~al.,
\newblock ``{LanguageBind}: Extending video-language pretraining to {N}-modality by language-based semantic alignment,''
\newblock in {\em The International Conference on Learning Representations}, 2024.

\end{thebibliography}

\end{document}